\documentclass[conference]{IEEEtran}

\usepackage{cite}
\usepackage{amsmath,amssymb,amsfonts}
\usepackage{graphicx}
\usepackage{textcomp}
\usepackage{xcolor}

\begin{document}

\title{Distance-Misaligned Training in Graph Transformers and Adaptive Graph-Aware Control}

\author{
    \IEEEauthorblockN{Qinhan Hou}
    \IEEEauthorblockA{\textit{University of Helsinki} \\
        Helsinki, Finland \\
        hou.qinhan@helsinki.fi}
    \and
    \IEEEauthorblockN{Jing Tang}
    \IEEEauthorblockA{\textit{University of Helsinki} \\
        Helsinki, Finland \\
        jing.tang@helsinki.fi}
}

\maketitle

\begin{abstract}
    Graph Transformers can mix information globally, but this flexibility also creates failure modes: some tasks require long-range communication while others are better served by local interaction. We study this through a synthetic node-classification benchmark on contextual stochastic block model graphs, where labels are generated by a controllable mixture of local and far-shell signals. We define distance-misaligned training as a mismatch between where label-relevant information lies and where the model allocates communication over graph distance. On this benchmark, we find three points. First, the preferred graph-distance bias changes systematically with task locality. Second, an oracle adaptive controller, given offline access to the task-side distance target, nearly matches the best fixed bias across regimes and strongly improves over a neutral baseline on mixed and local tasks. Third, a task-agnostic zero-gap controller is weaker, indicating that adaptation alone is not enough and that the control target matters. These results suggest that distance-resolved diagnosis is useful for understanding Graph Transformer failures and for designing graph-aware control.
\end{abstract}


\section{Introduction}
Graph Transformers have emerged as a competitive family of graph learners, but
their performance depends strongly on how structural bias is injected into
dense attention
~\cite{Ying2021DoTransformersReally,Rampasek2022RecipeGeneralPowerfulAdvancesinNeuralInformationProcessingSystems,Li2024Whatimprovesgeneralization}.
Recent analyses also suggest that unrestricted graph attention can create
undesirable regimes such as over-globalizing, over-aggregating, and
attention-based oversmoothing
~\cite{2024LessisMore,sun2025relieving,zhuo2025cross_aggregation,NEURIPS2023_6e4cdfdd,Dovonon2024SettingRecordStraight}.
These observations motivate a training-dynamics question: when should a model
communicate broadly across the graph, and when should it remain local?

We study this question through the lens of \emph{distance misalignment}. The
key idea is to separate two distributions over graph distance: task dependence,
which captures where label-relevant information lives, and model utilization,
which captures where the trained model allocates communication. A run is
\emph{under-reaching} when the task depends on larger graph distances but the
model stays too local, and \emph{over-globalizing} when the task is mostly
local but the model allocates too much mass to distant nodes. Our viewpoint is
complementary to prior work on Graph Transformer architectural bias: instead of
proposing a new attention mechanism, we ask whether training itself can be
diagnosed and steered in graph-distance space.

Our contribution is empirical. On a controlled synthetic
benchmark with explicit local and far-range supervision structure, we show that
distance-resolved mismatch diagnoses Graph Transformer failures and that a
simple graph-aware intervention can steer training.

\begin{figure*}[!t]
    \centering
    \includegraphics[width=\textwidth]{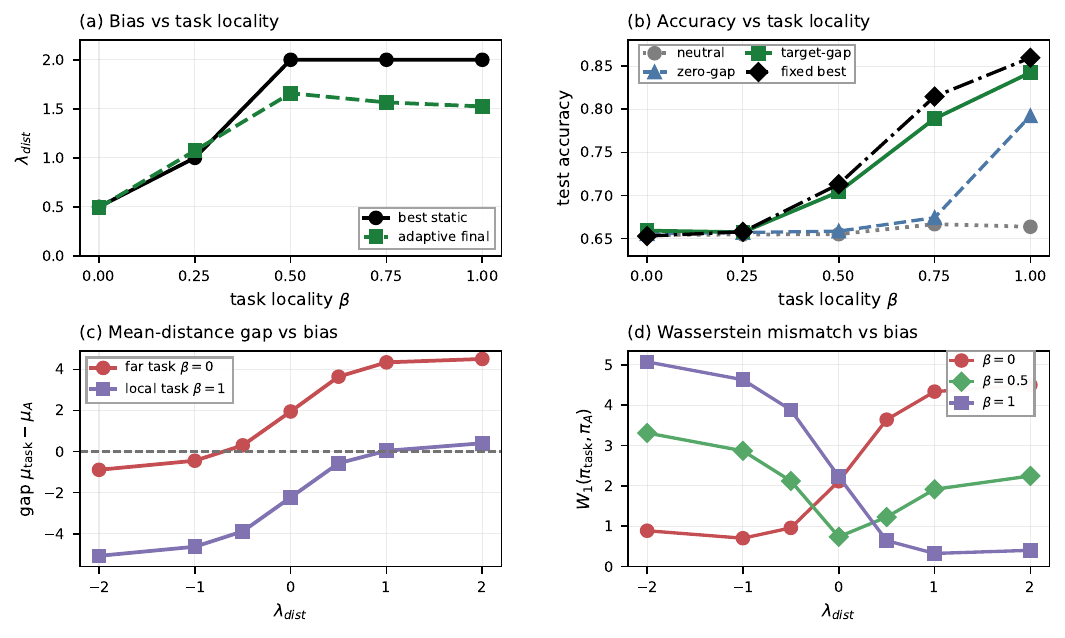}
    \caption{Results on the synthetic CSBM benchmark. (a) Validation-selected best fixed bias and final target-gap bias shift toward larger positive $\lambda_{dist}$ as tasks become more local. (b) Test accuracy of neutral, zero-gap, target-gap, and best fixed settings; zero-gap targets $\mu_{\mathrm{task}}-\mu_A=0$, while target-gap targets the oracle gap of the best fixed sweep. (c) Mean-distance gap shows how the same bias knob traverses over-globalizing and under-reaching regimes. (d) Wasserstein mismatch shows the same locality-dependent shift without sign.}
    \label{fig:summary}
\end{figure*}

\section{Experimental Setup}
We use a dense node-level Graph Transformer with a graph-distance bias added to
the attention logits:
\begin{equation}
    \mathrm{logit}_{ij} =
    \frac{q_i^\top k_j}{\sqrt{d}}
    +
    \lambda_{\mathrm{dist}} \, b_{\mathrm{dist}}(\mathrm{spd}(i,j)),
\end{equation}
where $\mathrm{spd}(i,j)$ is shortest-path distance and $b_{\mathrm{dist}}(r)=-r$; larger $\lambda_{\mathrm{dist}}$ favors local communication, while smaller (or negative) values favor more global communication.

The benchmark uses contextual stochastic block model graphs and a two-signal
node-classification task. Let $z_j$ be the latent node signal, $N_{\leq 1}(i)$
the 0--1 hop neighborhood of node $i$, and $S_{r^\star}(i)$ the shell at
distance $r^\star$. We define
\begin{equation}
    \begin{aligned}
        g_{\mathrm{loc}}(i) & = \frac{1}{|N_{\leq 1}(i)|} \sum_{j \in N_{\leq 1}(i)} z_j,   \\
        g_{\mathrm{far}}(i) & = \frac{1}{|S_{r^\star}(i)|} \sum_{j \in S_{r^\star}(i)} z_j,
    \end{aligned}
\end{equation}
standardize these scores over valid nodes to obtain $\hat g_{\mathrm{loc}}$ and $\hat g_{\mathrm{far}}$, and form labels by
\begin{equation}
    s(i)=\beta \hat g_{\mathrm{loc}}(i)+(1-\beta)\hat g_{\mathrm{far}}(i), \qquad
    y_i=\mathbf{1}[s(i)>0].
\end{equation}
The mixture coefficient $\beta \in [0,1]$ controls locality: $\beta=1$ is local and $\beta=0$ is far/global.

To measure training regime, we compute a task-side distance profile
$\pi_{\mathrm{task}}(r)$ and a shell-size-corrected attention profile
$\pi_A(r)$. We summarize mismatch by the mean-distance gap
$\mu_{\mathrm{task}}-\mu_A$, where
$\mu_{\mathrm{task}}=\sum_r r\,\pi_{\mathrm{task}}(r)$ and
$\mu_A=\sum_r r\,\pi_A(r)$, and by the Wasserstein-1 distance
$W_1(\pi_{\mathrm{task}},\pi_A)$. We compare neutral training
($\lambda_{\mathrm{dist}}=0$), fixed-\(\lambda_{\mathrm{dist}}\) sweeps, a
zero-gap controller that drives $\mu_{\mathrm{task}}-\mu_A$ toward $0$, and an
oracle target-gap controller, where \emph{oracle} means that the target gap is
supplied offline from the validation-selected best fixed
$\lambda_{\mathrm{dist}}$ for each $\beta$ rather than estimated from model-side
signals during training.
Positive gap indicates under-reaching and negative gap indicates over-globalizing.

\section{Results}
\subsection{Preferred graph-distance bias shifts with task locality}
Panel (a) of Fig.~\ref{fig:summary} shows a monotone shift toward larger
positive $\lambda_{\mathrm{dist}}$ as the task becomes more local. The
validation-selected bias is $0.5$ at $\beta=0$, $1.0$ at $\beta=0.25$, and
$2.0$ for $\beta \in \{0.5,0.75,1.0\}$. The far-task region is comparatively
flat, so the $\beta=0$ point should be read as near neutral; the adaptive
target-gap controller ends at similar final $\lambda_{\mathrm{dist}}$ values.

\subsection{Oracle target-gap control tracks the best fixed sweep}
Panel (b) of Fig.~\ref{fig:summary} gives the main performance comparison. The
neutral model is competitive only on the most global task; as locality
increases, its regret grows, while the oracle target-gap controller remains
close to the best fixed setting. For $\beta \in \{0.5,0.75,1.0\}$, target-gap
control reaches $0.704$, $0.789$, and $0.842$, compared with $0.655$, $0.667$,
and $0.664$ for fixed neutral, $0.659$, $0.674$, and $0.792$ for zero-gap, and
$0.713$, $0.815$, and $0.860$ for the best fixed runs. This indicates that
adaptation helps most when it targets a task-appropriate distance regime.

\subsection{Mismatch curves identify opposite failure modes}
Panels (c) and (d) of Fig.~\ref{fig:summary} provide the mechanistic
interpretation of the sweep. The mean-distance gap behaves differently for far
and local tasks as $\lambda_{\mathrm{dist}}$ varies, and the Wasserstein
mismatch follows the same locality-dependent shift in a sign-free form. For the
far task ($\beta=0$), increasing locality bias pushes the model from mild
over-globalizing behavior into strong under-reaching. For the local task
($\beta=1$), the trend is reversed: weak or negative bias is strongly
over-globalizing, while larger positive bias moves the model toward a small
positive gap near the best-performing region. This is the main mechanistic
evidence that the same control knob traverses distinct distance-misaligned
regimes. The Wasserstein panel shows the complementary summary on
representative far, mixed, and local tasks: its minimum shifts from weak or
negative bias toward larger positive bias as $\beta$ increases.

\section{Conclusion}
On this synthetic benchmark, distance-resolved mismatch provides a compact
diagnostic of whether a Graph Transformer is over-globalizing or under-reaching,
and graph-distance bias provides an effective control knob beyond aggregate
accuracy alone. The future work is to replace the oracle target with an
observable regime estimator built from model-side training signals and use it
for online scheduling that avoids poor training regimes. Moreover, the impact of the self-bias
term is still underexplored.

\clearpage
\section*{Acknowledgment}
This study was funded by the European Union (DTRIP4H, No. 101188432) and iCANDOC Precision
Cancer Medicine (PCM) pilot program from the Research Council of Finland.
\bibliographystyle{unsrt}
\bibliography{lib}

\end{document}